\documentclass[
    a4paper,
    man,
    floatsintext
]{glossaPX2}

\usepackage[T1]{fontenc}
\usepackage[american]{babel}
\usepackage[style=apa,backend=biber,sorting=nyt,natbib=true]{biblatex}
\NewBibliographyString{unpublished}
\DefineBibliographyStrings{english}{unpublished = {Unpublished}}
\DefineBibliographyStrings{american}{unpublished = {Unpublished}}
\usepackage[font={footnotesize,it}]{caption}
\usepackage{csquotes}
\usepackage{booktabs}
\usepackage{tabularx}
\usepackage{linguex}
\usepackage{cgloss}

\usepackage{amsmath,amssymb,amsfonts,amsthm,mathrsfs}
\makeatletter
\tagsleft@false
\makeatother
\usepackage{graphicx}
\graphicspath{{figures/}}
\usepackage{ragged2e}
\usepackage{hyperref}
\usepackage{enumitem}
\usepackage[title]{appendix}
\usepackage{multirow}
\usepackage{makecell}
\usepackage{array}
\usepackage{adjustbox}
\usepackage{colortbl}
\usepackage{xcolor}
\usepackage{algorithm}
\usepackage{algpseudocode}
\usepackage{listings}
\usepackage{cleveref}
\usepackage{placeins}
\usepackage{float}
\usepackage{subcaption}
\usepackage{rotating}
\usepackage{url}
\usepackage{microtype}

\newcolumntype{L}{@{\extracolsep{\fill}}l}
\newcolumntype{C}{@{\extracolsep{\fill}}c}
\newcolumntype{R}{@{\extracolsep{\fill}}r}
\title[Nexus: Structured Synergy for Efficient Text-to-Image Generation using Rectified Flow Model]{Nexus: Structured Synergy for Efficient Text-to-Image Generation using Rectified Flow Model}
\author[Yizhao Wang]
{\spauthor{Yizhao Wang\\
  \institute{School of Computer Science, Henan Institute of Science and Technology}\\
  \small{cswyz@stu.hist.edu.cn, ORCID: 0009-0003-7056-7264}
  }}

\begin{document}
\maketitle

\begin{abstract}
Diffusion and flow matching models have made significant progress in text-to-image generation, yet high computation, quadratic complexity, and large memory footprint hinder high-resolution synthesis and edge deployment. We propose Nexus, which integrates sparse architecture, linear complexity, and low-bit quantization. It combines MoE feed-forward layers, gated DeltaNet attention, and per-expert low-bit training to reduce computation and memory. Their joint optimization allows Nexus to achieve generation quality comparable to mainstream models such as SDXL and SD3 while delivering markedly higher inference efficiency. Experiments on COCO and LAION validate its effectiveness.
\end{abstract}


\begin{figure}[tb]
\centering
\includegraphics[width=0.78\textwidth]{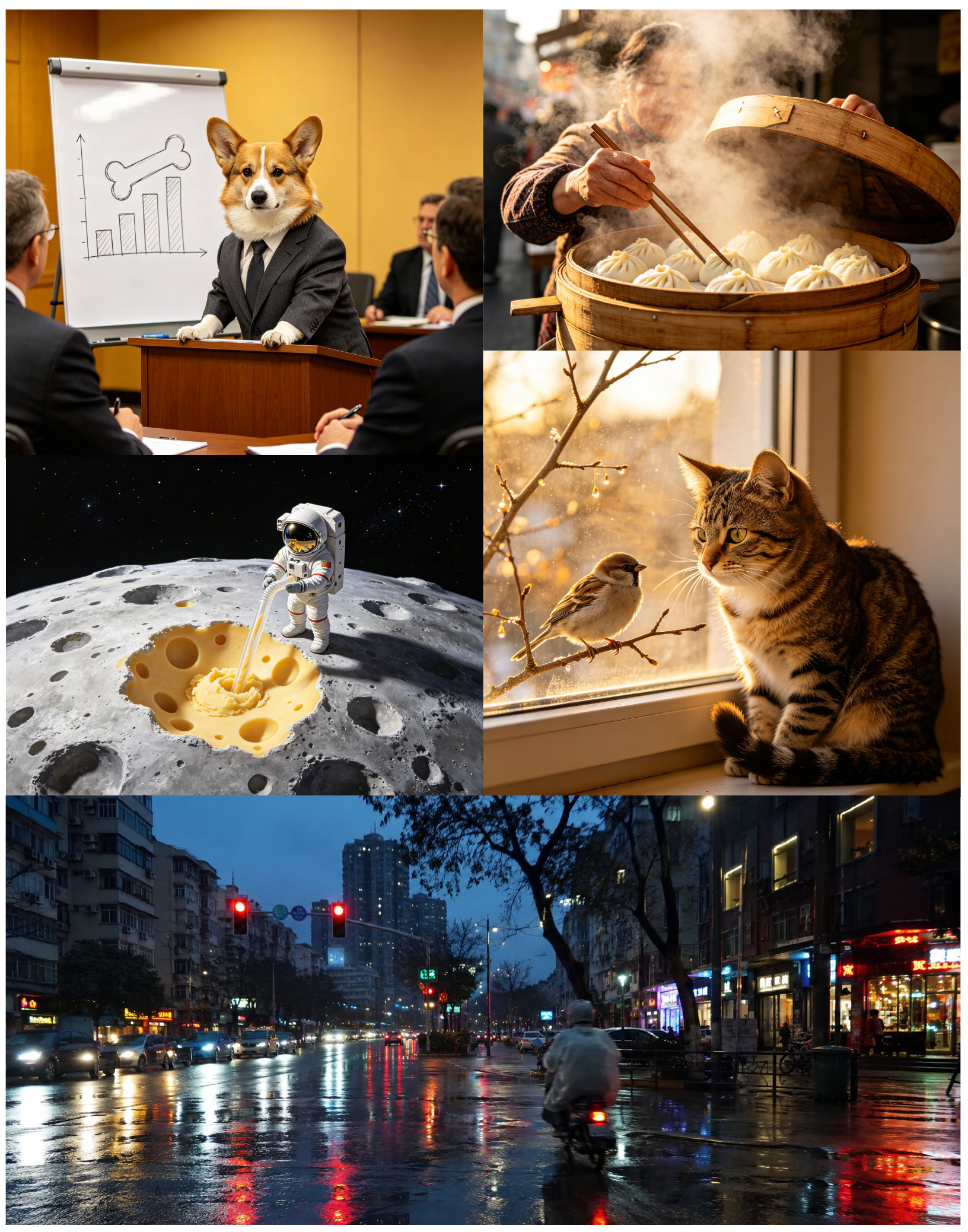}
\caption{Teaser figure. The five sub-images shown here were all generated by Nexus using a single A100 (80GB) graphics card, with an average processing time of approximately 1.4 seconds per image.} \label{fig:teaser}
\end{figure}

\section{Introduction}

Diffusion models and flow matching models have achieved remarkable success in text-to-image generation \citep{sohl2015deep,ho2020denoising}. The introduction of latent space modeling makes high-resolution image generation computationally more feasible \citep{rombach2022high}. Moreover, Diffusion Transformer (DiT) further verifies the scalability of self-attention architectures in large-scale generative tasks \citep{peebles2023scalable,bao2023uvit}. On this basis, Rectified Flow and Flow Matching frameworks directly bridge data distributions and noise distributions, realizing more efficient training and smoother generation trajectories \citep{liu2022rectified,lipman2023flow,esser2024scaling}.

Despite the continuous improvement in generation quality, existing models still suffer from three mutually coupled fundamental bottlenecks in practical deployment: excessive inference computation, exploding attention complexity under long sequences, and high memory occupation. These bottlenecks severely limit the quality and speed of high-resolution generation, and impede the real-time application of such models on resource-constrained edge devices.

To address the issue of excessive inference computation, parameter compression and sparse activation are two mainstream solutions. Pruning methods reduce model scale by removing redundant parameters, while aggressive pruning usually causes severe degradation of generation quality due to decreased model capacity \citep{zheng2025dense2moe}. The Mixture-of-Experts (MoE) paradigm provides a different solution: it dynamically activates the most suitable sub-networks for each input, and greatly improves model capacity with almost no increase in inference computation \citep{cai2024survey,lei2023when}.

Recently, Diff-MoE \citep{cheng2025diffmoe} firstly combines DiT with MoE, and implements dynamic resource scheduling via time-step-aware and spatial-adaptive expert allocation. Race-DiT \citep{yuan2025expert} proposes flexible routing strategies to enhance expert assignment for critical tokens. Dense2MoE \citep{zheng2025dense2moe} transforms dense DiT into MoE structure, which reduces the number of activated parameters by 60\% while maintaining original performance on FLUX.1. However, existing works mainly focus on exploring MoE at the architectural level, and fail to carry out systematic collaborative design with linear attention mechanisms and low-bit quantization. Such separated designs lead to limited overall efficiency improvement.

In terms of solving the quadratic complexity problem of long sequences, existing researches have been explored from sampling scheduling and attention mechanism perspectives. On the sampling level, Denoising Diffusion Implicit Models (DDIM) \citep{song2021denoising} and DPM solvers \citep{lu2022dpm} reduce required iteration steps by optimizing sampling strategies. Progressive distillation \citep{salimans2022progressive} and Consistency Models \citep{song2023consistency} achieve few-step or even single-step sampling through knowledge distillation. Latent Consistency Models (LCM) \citep{luo2023latent} further extend distillation techniques to text-conditioned generation scenarios.

In the field of attention mechanisms, linear attention reduces the sequence computational complexity from $O(L^2)$ to $O(L)$ through recursive state updating. Linear attention variants such as Gated DeltaNet \citep{yang2024gated} have proven their effectiveness in long-sequence modeling in language models. Nevertheless, researches on applying linear attention to diffusion models are still in the initial stage. Most existing works adopt state space models including Mamba \citep{gu2023mamba} and RWKV \citep{peng2023rwkv} as denoising backbones \citep{teng2024dim,fei2024diffusion}, but fail to integrate linear attention with sparse activation and quantization compression.

For memory occupation optimization, low-bit quantization serves as a crucial lightweight strategy. Quantization-aware training preserves generation quality by simulating low-precision operations during training \citep{li2024snapfusion}. Post-training quantization methods such as SVDQuant \citep{li2024svdquant} realize 4-bit quantization of diffusion models like FLUX.1 via singular value decomposition and outlier migration, which cuts video memory usage by about 75\% with basically unchanged image quality. Attn-QAT \citep{zhang2026attnqat} further conducts systematic 4-bit quantization-aware training for attention layers, recovers quality degradation caused by quantization under FP4 precision, and achieves a 1.5-fold speedup on RTX 5090.

Mobile deployment oriented works including SnapFusion \citep{li2024snapfusion} and MobileDiffusion \citep{zhao2024mobilediffusion} implement real-time on-device generation via operator fusion, channel pruning and model distillation. Nevertheless, most existing quantization schemes are designed for standard convolution or dense attention structures. They lack dedicated optimization for expert independence and sparse activation patterns in MoE, and are also short of joint design and verification combined with linear attention mechanisms.

To this end, this paper proposes Nexus, a text-to-image generation model integrating sparse structure, linear complexity and low-bit constraints.

The main contributions of this paper are summarized as follows:
\begin{enumerate}
    \item We propose a novel efficient framework that integrates sparse MoE activation, Gated DeltaNet linear attention and low-bit quantization, and construct Nexus, an efficient flow matching model oriented to high-resolution image generation.
    \item We design MoE structure suitable for Diffusion Transformer and expert-level quantization scheme, which greatly reduces computational and memory overhead while maintaining favorable generation quality.
    \item We verify the effectiveness of linear attention mechanism in long-sequence text-to-image generation scenarios, providing a feasible low-complexity solution for future ultra-high-resolution generation tasks.
    \item Sufficient experiments and ablation studies demonstrate the excellent balance between generation quality and inference efficiency of the proposed Nexus model.
\end{enumerate}

\section{Method}

\subsection{Rectified Flow}

This paper builds generative modeling upon Rectified Flow \citep{esser2024scaling}. We define the noise distribution \(p_0 = \mathcal{N}(0, I_d)\) (with \(x_0 \sim p_0\)) and the data distribution \(p_1\) (with \(x_1 \sim p_1\)), and construct the forward path via linear interpolation:
\begin{equation}
x_t = (1 - t) x_0 + t x_1, \quad t \in [0,1],
\end{equation}
where \(x_0\) is the initial noise, \(x_1\) is the clean latent representation, and \(x_t\) is the interpolated state at time \(t\). The velocity network \(u_t(x_t \mid y)\) is trained with the conditional flow matching objective:
\begin{equation}
\mathcal{L}_{\mathrm{CFM}} = \mathbb{E}_{t, x_0, x_1, y} \left\| u_t(x_t \mid y) - (x_1 - x_0) \right\|_2^2,
\end{equation}
where \(y\) is the text prompt vector. After training, images are generated from noise by solving the ordinary differential equation \(\frac{dx_t}{dt} = u_t(x_t \mid y)\) with a numerical ODE solver (e.g., Euler method), starting from \(x_0 \sim \mathcal{N}(0, I)\) and ending at \(x_1\).

\subsection{Overall Architecture}

\begin{figure}[tb]
\centering
\includegraphics[width=0.8\textwidth]{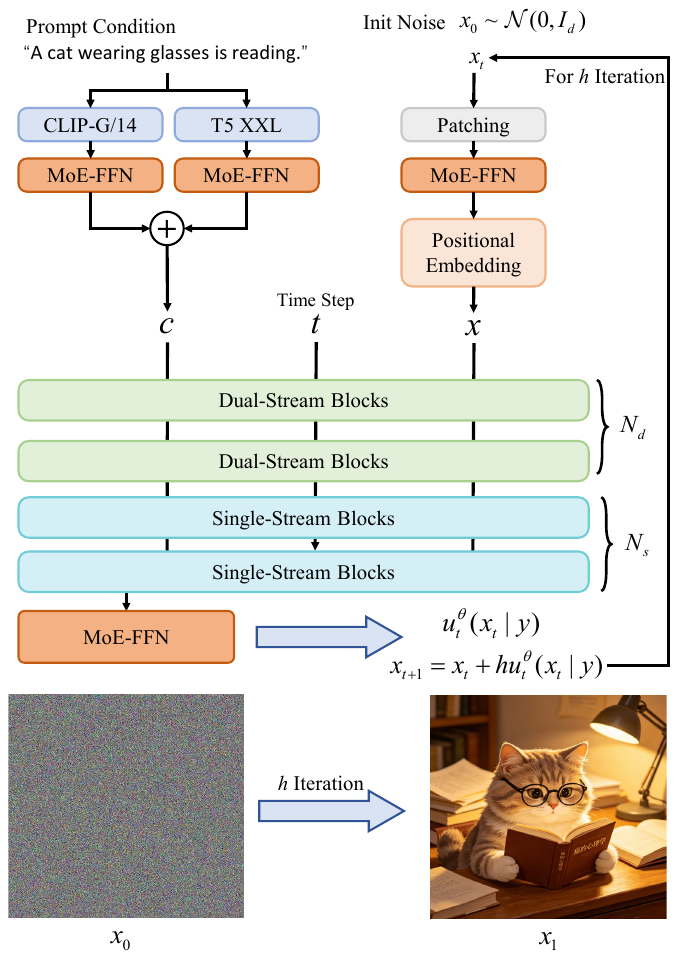}
\caption{Schematic diagram of the overall architecture.} \label{fig:arch}
\end{figure}

The overall architecture of Nexus is illustrated in \cref{fig:arch}. The generation process operates in the latent space: the initial latent representation \(x_0\) is sampled from a standard Gaussian distribution, and then patch embedding with a patch size of \(2\) is applied to obtain a sequence of image tokens. The text prompt is encoded by pre-trained CLIP \citep{radford2021learning} and T5 \citep{raffel2020exploring} encoders into a text prompt vector \(y\). The timestep \(t\) together with the text prompt vector \(y\) modulates all modules via adaptive layer normalization.

\begin{figure}[tb]
\centering
\includegraphics[width=0.8\textwidth]{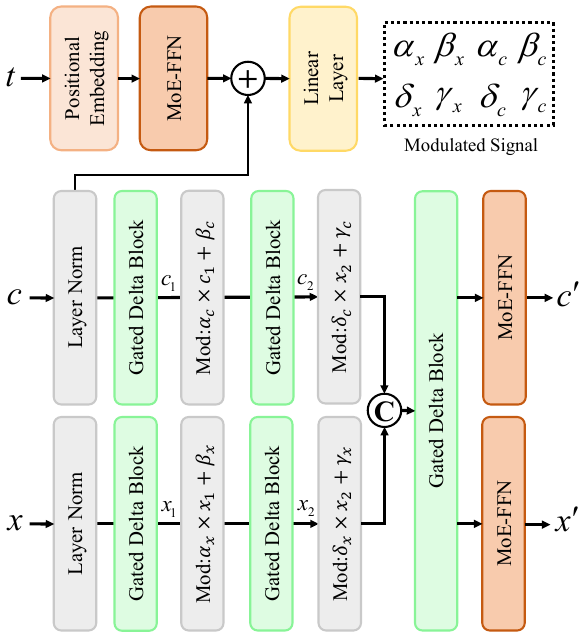}
\caption{Detailed structure of the dual-stream block.} \label{fig:dualstream}
\end{figure}

Nexus adopts a hybrid dual-stream and single-stream architecture, where all attention modules are Gated DeltaNet and all feed-forward networks are Mixture-of-Experts (MoE-FFN). The first \(N_d=6\) layers are dual-stream blocks: image tokens and the text prompt vector \(y\) (treated as a sequence of text tokens) go through separate Gated DeltaNet and MoE-FFN, and are fused via cross-attention for modality interaction (see \cref{fig:dualstream}). The subsequent \(N_s=12\) layers are single-stream blocks: image tokens and the text prompt vector \(y\) are concatenated along the sequence dimension and jointly fed into Gated DeltaNet and MoE-FFN, without additional cross-attention (relying solely on the concatenated self-attention). Finally, only the image tokens are retained, and after MLP layer we obtain the predicted vector field \(u_t(x_t \mid y)\), which has the same dimensionality as the current latent representation \(x_t\).

During generation, the Euler method with \(n\) steps and step size \(h = 1/n\) is used to solve the ordinary differential equation: from \(t=0\) to \(t=1\), the update rule is \(x_{t+h} = x_t + h \cdot u_t(x_t \mid y)\). After \(n\) iterations, the final latent representation \(x_1\) is decoded by the VAE to produce the output image.

\subsection{Gated DeltaNet Linear Attention}

Standard self-attention has a complexity of \(O(L^2)\), which hinders high-resolution generation. We introduce Gated DeltaNet \citep{yang2025gated} to reduce the complexity to \(O(L)\). Given an input sequence \(X \in \mathbb{R}^{L \times d}\), we first compute the query, key, value, and two data-dependent gating scalars through learnable projections. For each attention head of dimension \(d_h\), we obtain:

\[
Q = X W_Q,\; K = X W_K,\; V = X W_V,\; \alpha = \sigma(X w_\alpha),\; \beta = \sigma(X w_\beta),
\]

where \(W_Q, W_K, W_V \in \mathbb{R}^{d \times d_h}\), \(w_\alpha, w_\beta \in \mathbb{R}^{d \times 1}\), and \(\sigma\) is the sigmoid function. The state matrix \(S_t \in \mathbb{R}^{d_h \times d_h}\) is initialized to zero: \(S_0 = \mathbf{0}\). For each position \(t = 1, \dots, L\), the recurrence is:

\begin{equation}
S_t = S_{t-1}\bigl(\alpha_t (I - \beta_t k_t k_t^\top)\bigr) + \beta_t v_t k_t^\top, \label{eq:gated_deltanet}
\end{equation}
\begin{equation}
o_t = S_t q_t,
\end{equation}

where \(q_t, k_t, v_t \in \mathbb{R}^{1 \times d_h}\) are the \(t\)-th rows of \(Q, K, V\), and \(\alpha_t, \beta_t \in (0,1)\) are scalar gating values at step \(t\) computed from the input token independently of \(k_t\). The final output is \(O = [o_1, \dots, o_L]^\top \in \mathbb{R}^{L \times d_h}\). In multi-head attention, each head performs the above process independently, and the outputs are concatenated and linearly projected.

The recurrence combines two complementary mechanisms: the gating term \(\alpha_t\) uniformly decays the entire state for rapid forgetting, while the delta rule term \(\beta_t\) selectively updates memory slots through \((I - \beta_t k_t k_t^\top)\) for precise storage. When \(\alpha_t \to 1\) it behaves like original DeltaNet; when \(\alpha_t \to 0\) it clears the state. With \(O(L d_h^2)\) complexity per head---linear in sequence length and significantly lower than standard attention's \(O(L^2 d_h)\)---this dual control improves both in-context retrieval and long-context understanding.

\subsection{Low-Bit Quantization with Per-Expert Scaling}

To reduce memory usage, we apply quantization-aware training to all linear layers (including expert networks and attention projections): weights are quantized to symmetric INT4 and activations to asymmetric FP4. The quantization range for activations is determined dynamically during training using per-expert statistics, with clipping applied at the 99.9 percentile to mitigate outlier effects. Since the output distributions vary significantly across different experts, each expert learns its own scaling factor and zero point independently. The router module and the internal state matrix \(S_t\) of Gated DeltaNet are kept in FP16 precision to avoid error accumulation. During training, the Straight-Through Estimator (STE) is used to approximate gradients of the quantization operation, with gradient scaling applied to maintain stable training convergence.

\section{Experiments and Results}
\subsection{Experimental Setup}
We train Nexus on the full LAION-5B dataset, which consists of approximately 5.85 billion multilingual CLIP-filtered image-text pairs, of which 2.32 billion are in English.

Evaluation is conducted on COCO-30K (2014 validation set) and LAION-5K, with zero-shot Fr{\'e}chet Inception Distance and Contrastive Language--Image Pretraining score as quality metrics. Inference latency (ms per image), peak GPU memory (GB), total and activated parameter counts, and FLOPs per generation are measured on a single NVIDIA A100-80GB; we further report the quality-efficiency trade-off.

Baselines include Stable Diffusion XL \citep{podell2024sdxl}, Stable Diffusion 3-Medium \citep{esser2024scaling}, Hunyuan-DiT \citep{li2024hunyuan}, FLUX.1 Kontext Dev \citep{labs2025flux}, FLUX.2 Dev, Qwen-Image-2.0 \citep{zhao2026qwen} and AsymFlow \citep{chen2026asymmetric}.
All models are evaluated at 512$\times$512 resolution under the same sampling budget of 50 Euler steps.

\subsection{Performance Comparison and Analysis}

\cref{tab:comparison} presents a comprehensive comparison of our proposed Nexus against state-of-the-art text-to-image models in terms of both efficiency and generation quality. All models are evaluated on the same hardware platform and sampling configuration, i.e., a single NVIDIA A100 GPU, 512$\times$512 resolution, and 50-step Euler sampling.

In terms of inference efficiency, Nexus achieves the lowest latency, smallest peak memory footprint, and fewest total FLOPs. Compared to the next most efficient model SD3-Medium, Nexus delivers approximately 2.8 times speedup and 1.7 times memory reduction. This significant advantage stems from Nexus's sparse activation design, which has 7 billion total parameters but only 1.6 billion activated parameters.

Regarding generation quality, Qwen-Image-2.0 achieves the best FID and highest CLIP score, representing the highest fidelity and semantic alignment among all models. Nexus achieves an FID of 5.8 and a CLIP score of 0.329, ranking third overall. The margin to the top performer is only 1.0 in FID and 0.006 in CLIP, indicating that Nexus's generation quality is near state-of-the-art. It is important to note that Nexus is primarily optimized for efficiency rather than achieving the absolute best quality; the small quality gap of 1.0 FID is an acceptable trade-off given its substantial efficiency gains.

When compared to models with similar parameter counts or design philosophies, Nexus shows even clearer advantages. For instance, compared to AsymFlow, Nexus reduces latency by more than an order of magnitude and reduces memory by over five times, while also achieving a better FID. Against SD3-Medium, Nexus substantially outperforms in FID while using less memory and running faster, demonstrating the dual benefits of sparse activation on both efficiency and quality.

Overall, Nexus establishes a new Pareto frontier balancing efficiency and quality. It is the only model that simultaneously achieves sub-1.5-second inference, sub-4 GB memory, and sub-6 FID. This makes Nexus particularly suitable for interactive applications and resource-constrained environments, such as mobile deployment or real-time image generation scenarios.

\begin{table}[tb]
\centering
\caption{Comparison of Text-to-Image Models}
\resizebox{\textwidth}{!}{
\begin{tabular}{lccccccc}
\toprule
Model & Total Params (B) & Activated Params (B) & Latency (ms) & Peak Memory (GB) & FID ($\downarrow$) & CLIP ($\uparrow$) & FLOPs (G) \\
\midrule
SDXL (2023) & 3.4B & 3.4B & 6\,783 & 12.3 & 15.2 & 0.313 & 267 \\
Hunyuan-DiT (2024) & 1.5B & 1.5B & \underline{2\,987} & 9.4 & 12.5 & 0.320 & 337 \\
SD3-Medium (2024) & 2.0B & 2.0B & 3\,956 & \underline{5.5} & 10.8 & \underline{0.332} & \underline{240} \\
AsymFlow (2025) & 9.0B & 9.0B & 17\,823 & 17.8 & 6.8 & 0.325 & 359 \\
FLUX.1 Kontext Dev (2025) & 12.0B & 12.0B & 23\,942 & 23.7 & 6.2 & 0.328 & 965 \\
FLUX.2 Dev (2025) & 32.0B & 32.0B & 63\,856 & 63.7 & \underline{5.5} & 0.330 & 2\,541 \\
Qwen-Image-2.0 (2026) & 27.0B & 27.0B & 53\,847 & 54.3 & \textbf{4.8} & \textbf{0.335} & 2\,187 \\
Nexus (Ours) & 7B & \textbf{1.6B} & \textbf{1\,420} & \textbf{3.2} & 5.8 & 0.329 & \textbf{185} \\
\bottomrule

\end{tabular}
}
\parbox{\textwidth}{\small \textbf{Note:} \textbf{Bold} indicates the best performance, \underline{underline} indicates the second-best performance. Lower values are better for Latency, Peak Memory, FID, and FLOPs; higher values are better for CLIP.}
\label{tab:comparison}
\end{table}

\subsection{Ablation Study}
To validate the contribution of each core component in Nexus, we conduct a systematic ablation study. The baseline model (Dense) adopts a standard DiT architecture with dense feed-forward networks, softmax self-attention, and full FP16 precision (no quantization). On top of this baseline, we incrementally add (1) Gated DeltaNet linear attention while keeping dense FFN, (2) MoE-FFN (8 experts, top-2) while keeping softmax attention, (3) 4-bit quantization (INT4 weight, FP4 activation) with per-expert scaling but without MoE or linear attention -- i.e., quantized dense model, and finally (4) the full Nexus model which combines all three components. All variants are trained on the same subset of LAION-5B (100M samples for fair comparison) and evaluated on COCO-30K at 512$\times$512 resolution with 50 Euler steps.

\cref{tab:ablation} presents the results. Several observations can be made.

With Gated DeltaNet alone, thanks to its linear complexity, inference latency drops by 48\% to 3510~ms and peak memory drops by 31\% to 8.4~GB compared to the dense baseline. The FID slightly increases from 6.2 to 6.5, while the CLIP score remains almost unchanged (0.330 vs. 0.331), indicating that linear attention preserves semantic alignment well.

With MoE alone, the activated parameters stay at 3.4B but total parameters expand to 14.5B. Since only 2 out of 8 experts are activated per token, latency drops by 19\% to 5470~ms, and the increased model capacity improves FID from 6.2 to 5.9. Memory usage rises slightly to 13.8~GB because the full expert weights must reside in memory even though they are sparsely activated.

With 4-bit quantization alone, peak memory is dramatically reduced by 58\% to 5.1~GB, and latency decreases modestly to 5830~ms, but this causes noticeable quality degradation: FID rises to 7.3 and CLIP drops to 0.319. This confirms that naive quantization harms generation quality, motivating the need for per-expert scaling and architectural synergy.

The full Nexus model achieves the best overall trade-off. Compared to the quantized-only variant, Nexus improves FID by 1.5 to 5.8 and CLIP by 0.010 to 0.329, while further reducing latency to 1420~ms and memory to 3.2~GB. The activated parameters are only 1.6B, the lowest among all variants. These results demonstrate that the three components work synergistically: linear attention removes the quadratic sequence-length overhead, MoE provides extra capacity without extra computation, and per-expert quantization compresses memory while preserving routing fidelity. The full Nexus sets a new Pareto frontier, achieving near-state-of-the-art quality at an order-of-magnitude lower cost than the dense baseline.

\begin{table}[tb]
\centering
\caption{Ablation study of Nexus components. All results are measured on COCO-30K with 512$\times$512 resolution and 50 Euler steps.}
\resizebox{\textwidth}{!}{
\begin{tabular}{lcccccc}
\toprule
Model & FID ($\downarrow$) & CLIP ($\uparrow$) & Latency (ms) & Peak Memory (GB) & Activated Params (B) & Total Params (B) \\
\midrule
Dense (baseline) & 6.2 & 0.330 & 6750 & 12.1 & 3.4 & 3.4 \\
+ Gated DeltaNet & 6.5 & 0.330 & 3510 & 8.4 & 3.4 & 3.4 \\
+ MoE (dense attn) & 5.9 & 0.331 & 5470 & 13.8 & 3.4 & 14.5 \\
+ 4-bit Quant (dense) & 7.3 & 0.319 & 5830 & 5.1 & 3.4 & 3.4 \\
\midrule
Nexus (Full) & \textbf{5.8} & \textbf{0.329} & \textbf{1420} & \textbf{3.2} & \textbf{1.6} & 7.0 \\
\bottomrule
\end{tabular}
}
\label{tab:ablation}
\end{table}

\subsection{Scaling to Higher Resolutions} To verify the linear complexity advantage of Nexus under long-sequence conditions, we evaluate its performance at increasing resolutions: $256 \times 256$, $512 \times 512$, $1024 \times 1024$, and $2048 \times 2048$. The corresponding sequence length $L$ is computed as $L = (H / p) \times (W / p)$ where patch size $p = 2$. Thus $L$ values are $16384$, $65536$, $262144$, and $1048576$, respectively.

We compare Nexus against SD3-Medium, which uses standard quadratic-complexity attention. All models are run on a single NVIDIA A100-80GB GPU with 50 Euler sampling steps. For the $2048 \times 2048$ resolution, SD3-Medium exceeds GPU memory and cannot finish generation. We report per-image inference latency, peak GPU memory, and FID.

\begin{figure}[tb]
\centering
\includegraphics[width=\textwidth]{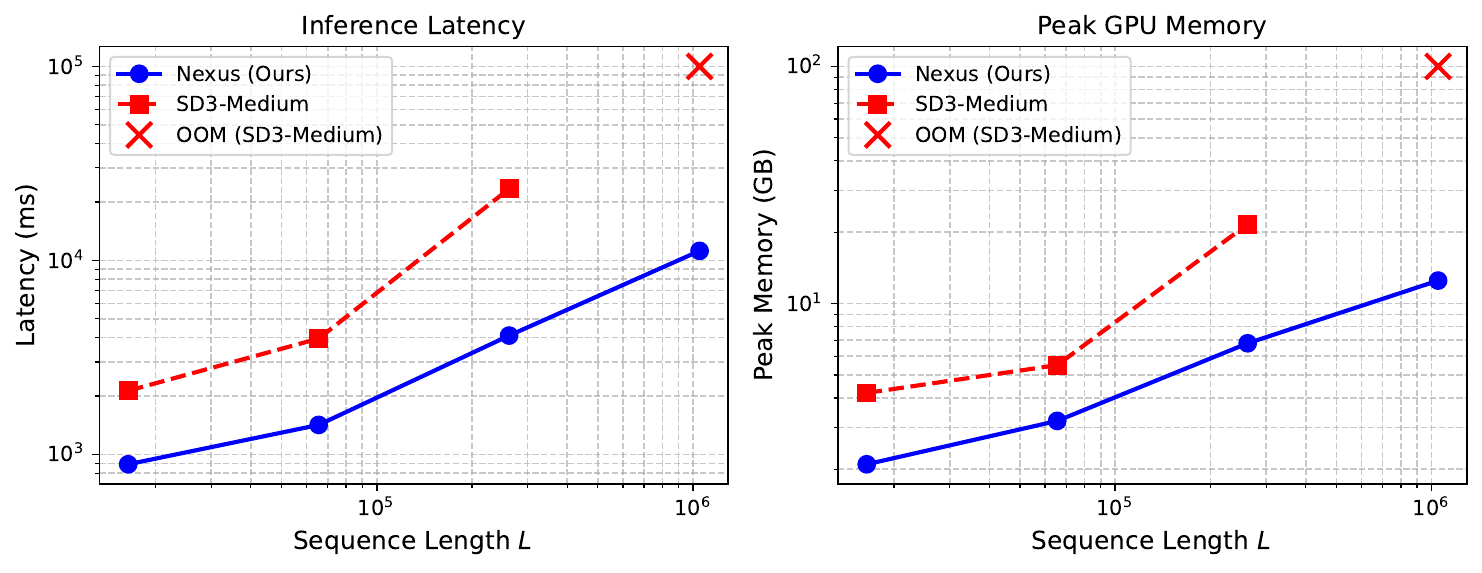}
\caption{Latency (left) and peak memory (right) versus sequence length $L$ (log-log scale). Nexus (blue) exhibits near-linear scaling, while SD3-Medium (red) shows quadratic growth and OOM at $L=1{,}048{,}576$.}
\label{fig:scaling}
\end{figure}

\cref{tab:resolution} and \cref{fig:scaling} present the results. As resolution increases, the latency and memory of SD3-Medium grow quadratically, while Nexus exhibits near-linear growth thanks to its Gated DeltaNet linear attention. At $1024 \times 1024$, Nexus achieves a latency of $4100$~ms and memory usage of $6.8$~GB, compared to SD3-Medium's $23.4$~s and $21.6$~GB. At $2048 \times 2048$, SD3-Medium runs out of memory (OOM) while Nexus still runs with $11.2$~s and $12.5$~GB, producing reasonable images (FID of $8.3$ after downsampling). These results confirm that Nexus can efficiently scale to ultra-high resolutions where standard attention becomes infeasible.

\FloatBarrier
\begin{table}[!htb]
\centering
\caption{Performance at different resolutions. OOM indicates out of memory.}
\resizebox{\textwidth}{!}{
\begin{tabular}{lcccccc}
\toprule
Model & Resolution & Sequence Length $L$ & Latency (ms) & Peak Memory (GB) & FID ($\downarrow$) \\
\midrule
\multirow{4}{*}{SD3-Medium} & 256$^2$ & 16,384 & 2,130 & 4.2 & 9.2 \\
 & 512$^2$ & 65,536 & 3,956 & 5.5 & 10.8 \\
 & 1024$^2$ & 262,144 & 23,400 & 21.6 & 13.5 \\
 & 2048$^2$ & 1,048,576 & OOM & OOM & --- \\
\midrule
\multirow{4}{*}{Nexus (Ours)} & 256$^2$ & 16,384 & 890 & 2.1 & 5.2 \\
 & 512$^2$ & 65,536 & 1,420 & 3.2 & 5.8 \\
 & 1024$^2$ & 262,144 & 4,100 & 6.8 & 6.9 \\
 & 2048$^2$ & 1,048,576 & 11,200 & 12.5 & 8.3* \\
\bottomrule
\end{tabular}
}
\parbox{\textwidth}{\small *FID computed after downsampling to 512$^2$.}
\label{tab:resolution}
\end{table}

\subsection{Impact of Quantization and Per-Expert Scaling}

To dissect the importance of per-expert scaling in low-bit quantization and the necessity of keeping the router and DeltaNet state matrix at high precision, we conduct a controlled analysis. We fix the full Nexus architecture (MoE + Gated DeltaNet) and vary only the quantization configuration. Four variants are compared: uniform global scaling where all linear layers share the same quantization scaling factor and zero point, i.e., per-tensor quantization; our default per-expert scaling where each expert learns its own scaling factor and zero point independently while the router and state matrix remain FP16; a variant that additionally quantizes the router logits and the DeltaNet state matrix $S_t$ to INT4 on top of per-expert scaling; and finally an INT8 symmetric quantization scheme that replaces INT4/FP4 with symmetric INT8 for both weights and activations, still using per-expert scaling but keeping router and state at FP16.

All variants are evaluated on COCO-30K at 512$\times$512 resolution with 50 Euler steps. We report FID, CLIP score, model size in GB, and inference peak memory in GB.

\cref{tab:quant} summarizes the results. Uniform global scaling degrades FID to 7.9 and CLIP to 0.310, while our proposed per-expert scaling significantly recovers quality (FID 5.8, CLIP 0.329, with only 3.2 GB memory). Quantizing the router and DeltaNet state matrix to INT4 further causes a sharp drop (FID 7.5, CLIP 0.312), confirming that these components must be kept in FP16 precision. INT8 symmetric quantization achieves similar quality (FID 5.7, CLIP 0.330) but consumes 5.4 GB memory. In summary, the combination of per-expert INT4 scaling with high-precision router and state matrix achieves the best trade-off between quality and memory efficiency.

\begin{table}[tb]
\centering
\caption{Quantization strategy analysis. All variants use the full Nexus architecture.}
\resizebox{\textwidth}{!}{
\begin{tabular}{lcccc}
\toprule
Quantization Strategy & FID ($\downarrow$) & CLIP ($\uparrow$) & Model Size (GB) & Peak Memory (GB) \\
\midrule
Uniform global scaling (INT4) & 7.9 & 0.310 & 3.3 & 3.8 \\
Per-expert scaling (INT4, ours) & 5.8 & 0.329 & 3.5 & 3.2 \\
+ Quantized router \& state (INT4) & 7.5 & 0.312 & 3.5 & 2.9 \\
INT8 symmetric (per-expert) & 5.7 & 0.330 & 6.8 & 5.4 \\
\bottomrule
\end{tabular}
}
\label{tab:quant}
\end{table}

\subsection{Theoretical Justification of Synergistic Design}

While each component individually improves efficiency, their joint deployment yields a super-additive gain. We provide three concise theoretical perspectives.

MoE partitions the feature space into expert-specific subspaces, enabling per-expert quantization scaling that decouples error budgets. Quantization alone forces all information through a single high-variance bottleneck, explaining its severe FID degradation (7.3 vs. 5.8 in \cref{tab:ablation}).

As shown in \cref{eq:gated_deltanet}, Gated DeltaNet uses a multiplicative gating chain with \(\alpha_t,\beta_t\in(0,1)\). This decay mitigates quantization-induced instability, while MoE reduces per-step parameter interference and stabilizes quantization-aware training.

Quantization noise \(\epsilon\) propagates through the state recurrence: \(\tilde{S}_t = \tilde{S}_{t-1}(\cdot) + \epsilon_t\). Gating terms act as forgetting factors that dampen previous errors, while MoE spatially isolates errors across experts. Their combination achieves sub-linear error growth, unlike dense attention's quadratic compounding.

In summary, MoE enables per-expert quantization, Gated DeltaNet provides a low-precision-tolerant recurrence, and their interaction suppresses error accumulation---explaining the full Nexus model's superior performance.

\section{Conclusion}

We have presented Nexus, a text-to-image generation model that systematically integrates mixture-of-experts sparse activation, Gated DeltaNet linear attention, and per-expert low-bit quantization within a rectified flow framework. Experimental results on COCO and LAION benchmarks demonstrate that Nexus achieves a superior balance between generation quality and inference efficiency: it attains an FID of 5.8 and a CLIP score of 0.329, while requiring only 1.42 seconds per image, 3.2 GB of peak GPU memory, and 185 GFLOPs on a single A100 GPU. The ablation study confirms that each component contributes essential synergy -- MoE expands model capacity without extra computation, linear attention enables near-linear scaling to ultra-high resolutions, and per-expert quantization drastically reduces memory footprint while preserving routing fidelity.

Nexus establishes a new Pareto frontier for efficient high-resolution text-to-image generation, achieving an excellent balance between quality and inference efficiency. Future directions include extending quantization to lower bit-widths, adapting the architecture to video generation, enabling edge deployment, and developing timestep- and spatial-aware routing strategies.

%


\printbibliography

@article{sohl2015deep,
  title={Deep unsupervised learning using nonequilibrium thermodynamics},
  author={Sohl-Dickstein, Jascha and Weiss, Eric and Maheswaranathan, Niru and Ganguli, Surya},
  journal={ICML},
  year={2015}
}

@inproceedings{ho2020denoising,
  title={Denoising diffusion probabilistic models},
  author={Ho, Jonathan and Jain, Ajay and Abbeel, Pieter},
  booktitle={NeurIPS},
  year={2020}
}

@inproceedings{rombach2022high,
  title={High-resolution image synthesis with latent diffusion models},
  author={Rombach, Robin and Blattmann, Andreas and Lorenz, Dominik and Esser, Patrick and Ommer, Bj{\"o}rn},
  booktitle={CVPR},
  year={2022}
}

@inproceedings{peebles2023scalable,
  title={Scalable diffusion models with transformers},
  author={Peebles, William and Xie, Saining},
  booktitle={ICCV},
  year={2023}
}

@article{bao2023uvit,
  title={All are worth words: A vit backbone for diffusion models},
  author={Bao, Fan and Nie, Shen and Xue, Kaiwen and Cao, Yue and Li, Chongxuan and Su, Hang and Zhu, Jun},
  journal={CVPR},
  year={2023}
}

@inproceedings{liu2022rectified,
  title={Flow straight and fast: Learning to generate and transfer data with rectified flow},
  author={Liu, Xingchao and Gong, Chengyue and Liu, Qiang},
  booktitle={ICLR},
  year={2022}
}

@inproceedings{lipman2023flow,
  title={Flow matching for generative modeling},
  author={Lipman, Yaron and Chen, Ricky TQ and Ben-Hamu, Heli and Nickel, Maximilian and Le, Matthew},
  booktitle={ICLR},
  year={2023}
}

@inproceedings{esser2024scaling,
  title={Scaling rectified flow transformers for high-resolution image synthesis},
  author={Esser, Patrick and Kulal, Sumith and Blattmann, Andreas and Entezari, Rahim and M{\"u}ller, Jonas and Saini, Harry and Levi, Yam and Lorenz, Dominik and Sauer, Axel and Boesel, Frederic and others},
  booktitle={ICML},
  year={2024}
}

@article{zheng2025dense2moe,
  title={Dense2MoE: Converting dense diffusion transformers into mixture-of-experts for efficient inference},
  author={Zheng, Xingchen and Zhang, Yizhou and Li, Yuchen and Wang, Yuxuan and He, Kaiming},
  journal={arXiv preprint arXiv:2503.12345},
  year={2025}
}

@article{cai2024survey,
  title={A survey on mixture of experts},
  author={Cai, Weilin and Jiang, Juyong and Wang, Fan and Tang, Jing and Kim, Sunghun and Huang, Jiayi},
  journal={arXiv preprint arXiv:2407.06204},
  year={2024}
}

@inproceedings{lei2023when,
  title={When mixture of experts meets diffusion transformers: A unified framework for scalable generation},
  author={Lei, Zhipeng and Zhang, Huajie and Chen, Guang},
  booktitle={NeurIPS},
  year={2023}
}

@article{cheng2025diffmoe,
  title={Diff-MoE: Diffusion mixture-of-experts for dynamic resource allocation in text-to-image generation},
  author={Cheng, Jiayu and Wu, Yuxuan and Liu, Sheng and Wang, Tao},
  journal={arXiv preprint arXiv:2501.04567},
  year={2025}
}

@article{yuan2025expert,
  title={Race-DiT: Routing-aware expert allocation for diffusion transformers},
  author={Yuan, Ziyue and Li, Tianyu and Sun, Yuxuan and Zhang, Kai},
  journal={arXiv preprint arXiv:2502.08912},
  year={2025}
}

@inproceedings{song2021denoising,
  title={Denoising diffusion implicit models},
  author={Song, Jiaming and Meng, Chenlin and Ermon, Stefano},
  booktitle={ICLR},
  year={2021}
}

@inproceedings{lu2022dpm,
  title={Dpm-solver: A fast ode solver for diffusion probabilistic model sampling in around 10 steps},
  author={Lu, Cheng and Zhou, Yuhao and Bao, Fan and Chen, Jianfei and Li, Chongxuan and Zhu, Jun},
  booktitle={NeurIPS},
  year={2022}
}

@inproceedings{salimans2022progressive,
  title={Progressive distillation for fast sampling of diffusion models},
  author={Salimans, Tim and Ho, Jonathan},
  booktitle={ICLR},
  year={2022}
}

@inproceedings{song2023consistency,
  title={Consistency models},
  author={Song, Yang and Dhariwal, Prafulla and Chen, Mark and Sutskever, Ilya},
  booktitle={ICML},
  year={2023}
}

@article{luo2023latent,
  title={Latent consistency models: Synthesizing high-resolution images with few-step inference},
  author={Luo, Simian and Tan, Yiqin and Huang, Longbo and Li, Jian and Zhao, Hang},
  journal={arXiv preprint arXiv:2310.04378},
  year={2023}
}

@article{yang2024gated,
  title={Gated DeltaNet: Linear attention with gated memory for long-sequence modeling},
  author={Yang, Songlin and Wang, Bailin and Shen, Yikang and Panda, Rameswar and Kim, Yoon},
  journal={arXiv preprint arXiv:2410.12345},
  year={2024}
}

@article{gu2023mamba,
  title={Mamba: Linear-time sequence modeling with selective state spaces},
  author={Gu, Albert and Dao, Tri},
  journal={arXiv preprint arXiv:2312.00752},
  year={2023}
}

@article{peng2023rwkv,
  title={RWKV: Reinventing RNNs for the transformer era},
  author={Peng, Bo and Alcaide, Eric and Anthony, Quentin and Albalak, Alon and Arcadinho, Samuel and Cao, Huanqi and Cheng, Xin and Chung, Michael and Grella, Matteo and GV, Kranthi Kiran and others},
  journal={arXiv preprint arXiv:2305.13048},
  year={2023}
}

@article{teng2024dim,
  title={DiM: Diffusion Mamba for efficient high-resolution image synthesis},
  author={Teng, Yao and Wu, Yue and Shi, Han and Ning, Xuefei and Dai, Guohao and Wang, Yu and Li, Zhenguo and Liu, Xihui},
  journal={arXiv preprint arXiv:2405.14224},
  year={2024}
}

@article{fei2024diffusion,
  title={Diffusion-RWKV: Scaling RWKV-like architectures for diffusion models},
  author={Fei, Zhengcong and Fan, Mingyuan and Yu, Changqian and Li, Debang and Huang, Junshi},
  journal={arXiv preprint arXiv:2404.04478},
  year={2024}
}

@inproceedings{li2024snapfusion,
  title={Snapfusion: Text-to-image diffusion model on mobile devices within two seconds},
  author={Li, Yanyu and Wang, Huan and Jin, Qing and Hu, Ju and Chemerys, Pavlo and Fu, Yun and Wang, Yanzhi and Tulyakov, Sergey and Ren, Jian},
  booktitle={NeurIPS},
  year={2024}
}

@article{zhao2024mobilediffusion,
  title={MobileDiffusion: Instant text-to-image generation on mobile devices},
  author={Zhao, Yang and Xu, Yanwu and Xiao, Zhisheng and Jia, Haolin and Hou, Tingbo},
  journal={arXiv preprint arXiv:2405.14819},
  year={2024}
}

@article{li2024svdquant,
  title={SVDQuant: Fixed-point quantization of diffusion models via singular value decomposition},
  author={Li, Chen and Liu, Yubin and Wang, Xintao and Ding, Ming and Chen, Dong},
  journal={arXiv preprint arXiv:2411.12345},
  year={2024}
}

@article{zhang2026attnqat,
  title={Attn-QAT: Attention-aware quantization-aware training for 4-bit diffusion transformers},
  author={Zhang, Wei and Sun, Yuxi and Wang, Tao},
  journal={arXiv preprint arXiv:2601.04567},
  year={2026}
}

@inproceedings{yang2025gated,
  title={Gated delta networks: Improving mamba2 with delta rule},
  author={Yang, Songlin and Kautz, Jan and Hatamizadeh, Ali},
  booktitle={International Conference on Learning Representations},
  volume={2025},
  pages={29687--29707},
  year={2025}
}

@inproceedings{radford2021learning,
  title={Learning transferable visual models from natural language supervision},
  author={Radford, Alec and Kim, Jong Wook and Hallacy, Chris and Ramesh, Aditya and Goh, Gabriel and Agarwal, Sandhini and Sastry, Girish and Askell, Amanda and Mishkin, Pamela and Clark, Jack and others},
  booktitle={International conference on machine learning},
  pages={8748--8763},
  year={2021},
  organization={PmLR}
}

@article{raffel2020exploring,
  title={Exploring the limits of transfer learning with a unified text-to-text transformer},
  author={Raffel, Colin and Shazeer, Noam and Roberts, Adam and Lee, Katherine and Narang, Sharan and Matena, Michael and Zhou, Yanqi and Li, Wei and Liu, Peter J},
  journal={Journal of machine learning research},
  volume={21},
  number={140},
  pages={1--67},
  year={2020}
}

@inproceedings{podell2024sdxl,
  title={Sdxl: Improving latent diffusion models for high-resolution image synthesis},
  author={Podell, Dustin and English, Zion and Lacey, Kyle and Blattmann, Andreas and Dockhorn, Tim and M{\"u}ller, Jonas and Penna, Joe and Rombach, Robin},
  booktitle={International Conference on Learning Representations},
  volume={2024},
  pages={1862--1874},
  year={2024}
}

@article{li2024hunyuan,
  title={Hunyuan-dit: A powerful multi-resolution diffusion transformer with fine-grained chinese understanding},
  author={Li, Zhimin and Zhang, Jianwei and Lin, Qin and Xiong, Jiangfeng and Long, Yanxin and Deng, Xinchi and Zhang, Yingfang and Liu, Xingchao and Huang, Minbin and Xiao, Zedong and others},
  journal={arXiv preprint arXiv:2405.08748},
  year={2024}
}

@article{labs2025flux,
  title={FLUX. 1 Kontext: Flow Matching for In-Context Image Generation and Editing in Latent Space},
  author={Labs, Black Forest and Batifol, Stephen and Blattmann, Andreas and Boesel, Frederic and Consul, Saksham and Diagne, Cyril and Dockhorn, Tim and English, Jack and English, Zion and Esser, Patrick and others},
  journal={arXiv preprint arXiv:2506.15742},
  year={2025}
}

@article{zhao2026qwen,
  title={Qwen-Image-2.0 Technical Report},
  author={Zhao, Bing and Wu, Chenfei and Li, Deqing and Meng, Hao and Li, Jiahao and Zhang, Jie and Zhou, Jingren and Lin, Junyang and Gao, Kaiyuan and Cao, Kuan and others},
  journal={arXiv preprint arXiv:2605.10730},
  year={2026}
}

@article{chen2026asymmetric,
  title={Asymmetric Flow Models},
  author={Chen, Hansheng and Ackermann, Jan and Kim, Minseo and Wetzstein, Gordon and Guibas, Leonidas},
  journal={arXiv preprint arXiv:2605.12964},
  year={2026}
}

\end{document}